\documentclass[10pt,twocolumn,letterpaper]{article}

\usepackage[pagenumbers]{cvpr} % To force page numbers, e.g. for an arXiv version

\usepackage{graphicx}
\usepackage{amsmath}
\usepackage{float}
\usepackage{amssymb}
\usepackage{booktabs}
\usepackage{caption}
\usepackage{bm}

\DeclareMathOperator*{\argmin}{arg\,min}

\usepackage[pagebackref,breaklinks,colorlinks]{hyperref}

\usepackage[capitalize]{cleveref}
\crefname{section}{Sec.}{Secs.}
\Crefname{section}{Section}{Sections}
\Crefname{table}{Table}{Tables}
\crefname{table}{Tab.}{Tabs.}

\def\cvprPaperID{*****} % *** Enter the CVPR Paper ID here
\def\confName{CVPR}
\def\confYear{2022}

\begin{document}

\title{Anything-3D: Towards Single-view Anything Reconstruction in the Wild}

\author{Qiuhong Shen{\footnotemark[1]} \, \, \, \,
        Xingyi Yang{\footnotemark[1]} \, \,  \, \,
        Xinchao Wang {}
        \\
        National University of Singapore
        % \textsuperscript{2} SenseTime Research \\
        % \textsuperscript{3} The University of Sydney 
        % \textsuperscript{4} Peng Cheng Laboratory 
        % \textsuperscript{5} Shanghai AI Laboratory \\
        \\
        \tt\small{
            \{qiuhong.shen,xyang\}@u.nus.edu,
            xinchao@nus.edu.sg}
}

% \maketitle
\twocolumn[{
\maketitle
\begin{center}
    \captionsetup{type=figure}
    \resizebox{\linewidth}{!}{\includegraphics[width=\textwidth]{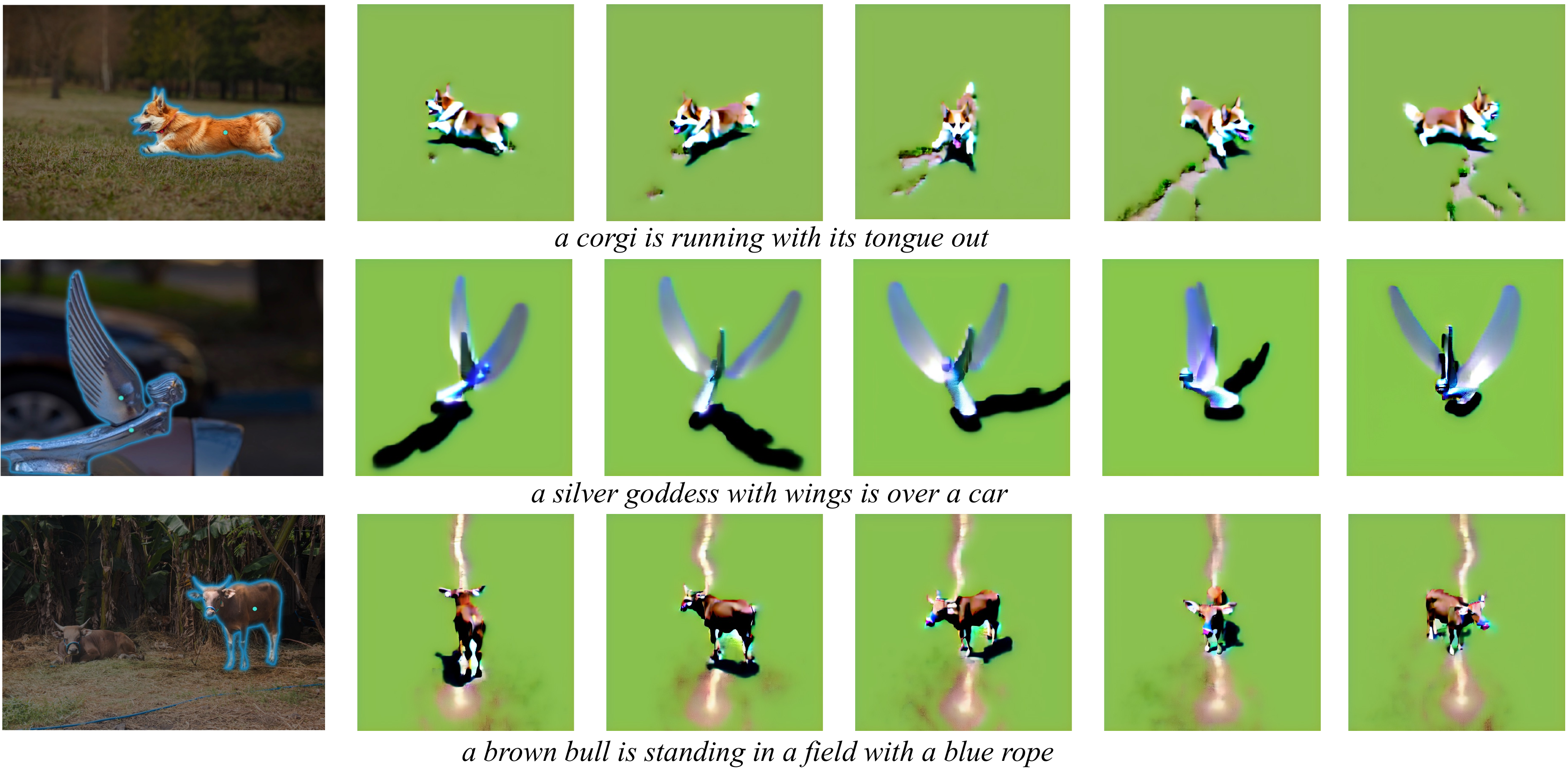}}
    \captionof{figure}{The Anything-3D framework proficiently recovers the 3D geometry and texture of any object from a single-view image captured in uncontrolled environments. Despite significant variations in camera perspective and object properties, our approach consistently delivers reliable recovery results.}
    \label{fig:demo_0}
\end{center}
}]
\renewcommand{\thefootnote}{\fnsymbol{footnote}}
\footnotetext[1]{Equal Contribution}
\footnotetext[2]{Work in progress}

%%%%%%%%% ABSTRACT
\begin{abstract}
   3D reconstruction from a single-RGB image in unconstrained real-world scenarios presents numerous challenges due to the inherent diversity and complexity of objects and environments. In this paper, we introduce Anything-3D, a methodical framework that ingeniously combines a series of visual-language models and the Segment-Anything object segmentation model to elevate objects to 3D, yielding a reliable and versatile system for single-view conditioned 3D reconstruction task. Our approach employs a BLIP model to generate textural descriptions, utilizes the Segment-Anything model for the effective extraction of objects of interest, and leverages a text-to-image diffusion model to lift object into a neural radiance field. Demonstrating its ability to produce accurate and detailed 3D reconstructions for a wide array of objects, \emph{Anything-3D\footnotemark[2]} shows promise in addressing the limitations of existing methodologies. Through comprehensive experiments and evaluations on various datasets, we showcase the merits of our approach, underscoring its potential to contribute meaningfully to the field of 3D reconstruction. 
   Demos and code will be available at \href{https://github.com/Anything-of-anything/Anything-3D}{https://github.com/Anything-of-anything/Anything-3D}.
\end{abstract}

% \renewcommand{\thefootnote}{\fnsymbol{footnote}}
% \footnotetext[1]{Equal Contribution.}

%%%%%%%%% BODY TEXT
\section{Introduction}
\label{sec:intro}

The quest to reconstruct 3D objects from 2D images is central to the evolution of computer vision and has far-reaching consequences for robotics, autonomous driving, augmented and virtual reality, as well as 3D printing. In spite of considerable progress in recent years, the task of single-image object reconstruction in unstructured settings continues to be an intellectually engaging and daunting problem.

The task at hand entails generating a 3D representation of one or more objects from a single 2D image, potentially encompassing point clouds, meshes, or volumetric representations. However, the problem is fundamentally \emph{ill-posed}. Due to the intrinsic ambiguity that arises from the 2D projection, it is impossible to unambiguously determine an object's 3D structure.

Reconstructing objects in the wild is primarily complicated by the vast diversity in shapes, sizes, textures, and appearances. Furthermore, objects in real-world images are frequently subject to partial occlusion, hindering the accurate reconstruction of obscured parts. Variables such as lighting and shadows can drastically affect object appearance, while differences in angles and distances can lead to significant alterations in 2D projections.

In this paper, we present Anything-3D, a pioneering and systematic framework that fuses visual-language models with object segmentation to effortlessly transform objects from 2D to 3D, culminating in a powerful and adaptable system for single-view reconstruction tasks. Our framework overcomes these challenges by incorporating the Segment-Anything model (SAM) alongside a series of visual-language foundation models to retrieve the 3D texture and geometry of a given image. Initially, a Bootstrapping Language-Image Pre-training (BLIP) estimates the textual description of the image. Next, SAM identifies the object region of interest. The segmented object and description are subsequently employed to execute the 3D reconstruction task. Our 3D reconstruction model leverages a pre-trained 2D text-to-image diffusion model to carry out image-to-3D synthesis. In particular, we utilize a text-to-image diffusion model that processes a 2D image and a textual description of an object, employing score distillation to train a neural radiance field specific to the image. Through rigorous experiments on diverse datasets, we showcase the effectiveness and adaptability of our approach, outstripping existing methods in terms of accuracy, robustness, and generalization capability. Moreover, we provide a thorough analysis of the challenges inherent in 3D object reconstruction in the wild and discuss how our framework addresses these obstacles.

By harmoniously melding the zero-shot vision and linguistic comprehension abilities of the foundation models, our framework excels at reconstructing an extensive range of objects from real-world images, producing precise and intricate 3D representations applicable to a myriad of use cases. Ultimately, our proposed framework, Anything-3D, represents a groundbreaking stride in the domain of 3D object reconstruction from single images.

\section{Related Works}

\noindent{\textbf{Single view 3D reconstruction.}}
A 3D object has details of geometry with texture and can be projected to 2D image at any views. Dense view images with camera poses or depth information are always required to rebuild 3D objects. Though the general 3D reconstruction task has made significant strides in past decades, exploration of reconstructing 3D objects from single views is still limited. While single-view reconstruction has been attempted for specific object categories, such as faces, human bodies, and vehicles, these approaches rely on strong category-wise priors like meshes and CAD models. The reconstruction of arbitrary 3D objects from single-view images remains largely unexplored and requires expert knowledge, often taking several hours.

 Reconstructing an arbitrary 3D object from a single-view image is inherently an ill-posed optimization problem, making it extremely challenging and largely unexplored. However, recent work has begun incorporating general visual knowledge into 3D reconstruction models, enabling the generation of 3D models from single-view images. NeuraLift360~\cite{lift360} employ pre-trained depth-estimator and 2D diffusion priors to recover coarse geometry and textures of 3D objects from single image. 3DFuse~\cite{3dfuse} initialize geometry with estimated point cloud~\cite{point_e}, then fine-grained geometry and textures are learned from single image with a 2D diffusion prior with score distillation~\cite{sjc}. Another work MCC~\cite{mcc} learns from object-centric videos to generate 3D objects from single RGB-D image. These works show promising results in this direction but still rely heavily on human-specified text or captured depth information.

Our work aims to reconstruct arbitrary objects of interest from single-perspective, in-the-wild images, which are ubiquitous and easily accessible in the real world. By tackling this challenging problem, we hope to unlock new potential for 3D object reconstruction from limited data sources.

% \vspace{2mm}
\noindent{\textbf{Text-to-3D Generative Models.}} An alternative approach to generating 3D objects involves using text prompts as input. Groundbreaking works like DreamFusion~\cite{dreamfusion} pioneered text-to-3D generative models using score distillation on pre-trained 2D generative models. This inspired a series of subsequent works, such as SJC~\cite{sjc}, Magic3D~\cite{magic3d}, and Fantasia3D~\cite{fantasia3d}, which focused on generating voxel-based radiance fields or high-resolution 3D objects from text using various optimization techniques and neural rendering methods. Point-E~\cite{point_e} represents another approach, generating point clouds from either images or text.

In contrast to these text-based methods, our work targets generating 3D objects from single-view images. We strive to ensure the generated 3D objects are as consistent as possible with the given single-view images, pushing the boundaries of 3D vision and machine learning research. Our approach demonstrates the potential for reconstructing complex, real-world objects from limited information, inspiring further exploration and innovation in the field.

\section{Preliminary}

\subsection{Problem definition}

Our problem refers to the task of automatically generating a 3D model of an object in a real-world image, using only a single 2D image as input, without any prior information about the object category.

Let $I \in \mathbb{R}^{H\times W\times3}$ be a real-world image represented as a matrix of RGB values, where $H$ and $W$ represent the height and width of the image, respectively. The goal is to generate a 3D model of the object in the image, represented by a NeRF function $C(\mathbf{x}; \theta)$, {where $\mathbf{x} \in \mathbb{R}^5$ represents a point and its view direction in 3D space,} 
% $\mathbf{p} = [\mathbf{c}^T, \mathbf{q}^T]^T \in \mathbb{R}^7$ represents the camera and view parameters, 
and $\theta$ represents the parameters of the neural network that predicts the RGB value $\mathbf{c}$ and volume density $\mathbf{\sigma}$ of the sampled location.

\subsection{Challenges}

The problem of single view, 3D reconstruction in the wild for arbitrary class image presents a thrilling but formidable challenge, with numerous complexities and constraints to overcome.

\begin{enumerate}
\item \textbf{Arbitrary Class.} Previous models for reconstructing 3D images succeed in specific object categories but often fail when applied to new or unseen categories without a parametric form, highlighting the need for new methods to extract meaningful features and patterns from images without prior assumptions.

\item \textbf{In the Wild.} Real-world images present various challenges, such as occlusions, lighting variations, and complex object shapes, which can affect the accuracy of 3D reconstruction. Overcoming these difficulties requires networks to account for these variations while inferring the 3D structure of the scene.

\item \textbf{No Supervision.} The absence of a large-scale dataset with paired single view images and 3D ground truth limits the training and evaluation of proposed models, hindering their ability to generalize to a wide range of object categories and poses. The lack of supervision also makes it difficult to validate the accuracy of the reconstructed 3D models.

\item \textbf{Single-View.} Inferring the correct 3D model from a single 2D image is inherently ill-posed, as there are infinitely many possible 3D models that could explain the same image. This leads to ambiguity in the problem and makes it challenging to achieve high accuracy in reconstructed 3D models, requiring networks to make assumptions and trade-offs to arrive at a plausible 3D structure.
\end{enumerate}

\begin{figure*}
    \centering
\includegraphics[width=\linewidth]{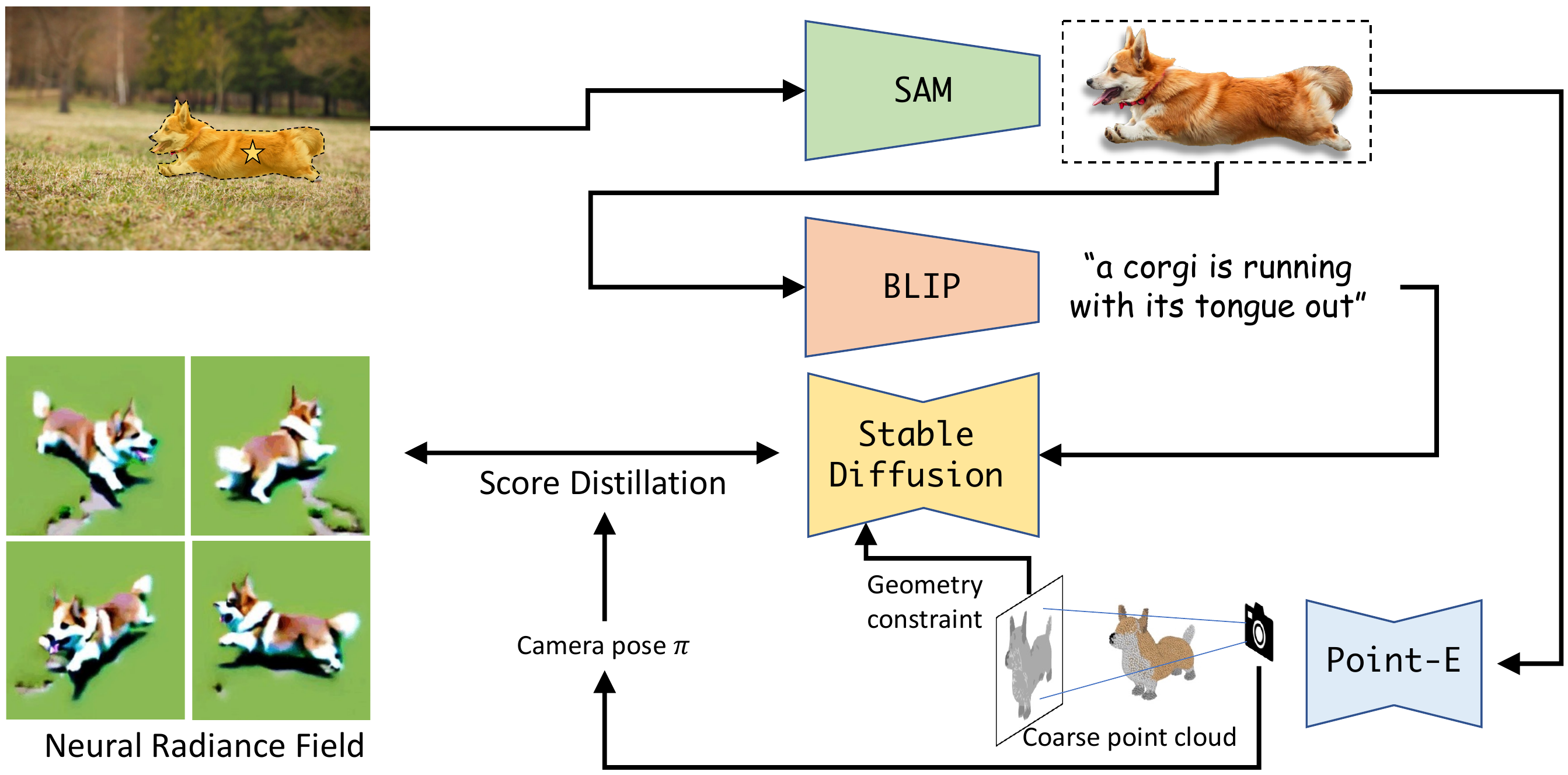}
    \caption{Anything-3D combines visual-language models and object segmentation for efficient single-view 3D reconstruction. The framework employs BLIP for textual description generation and SAM for object segmentation, followed by 3D reconstruction using a pre-trained 2D text-to-image diffusion model. This model processes the 2D image and textual description, utilizing score distillation to train a neural radiance field specific to the image for image-to-3D synthesis.}
    \label{fig:pipeline}
\end{figure*}

\subsection{What Do We Mean by ``Anything''?}

The term "anything" emphasizes the exceptional and ambitious nature of our approach. By using this term, we highlight the capability of our framework to generate 3D models from a single input image, irrespective of the object category or scene complexity. This aspect differentiates our approach from previous works that may be confined to certain object classes or restricted environments.

No prior work, to the best of our knowledge, has successfully tackled the task of Single-view Anything Reconstruction in the Wild, leaving the feasibility of a solution uncertain. In this paper, we boldly address this challenge by devising a unified framework that combines a series of foundational models, each boasting unparalleled generalization abilities. Our framework is designed to enable robust and accurate 3D object reconstruction from a single image, irrespective of object category or environmental complexities.

Our contribution specifically leverages state-of-the-art visual-language models to extract semantic information from the input image. This extracted information is then utilized to inform a high-precision object segmentation algorithm. Through the careful fusion of these components, our Single-view Anything Reconstruction in the Wild framework generates detailed 3D object reconstructions from a single input image, effectively overcoming challenges posed by complex and unstructured environments.

\section{Methodology}
\label{sec:methodology}

In this section, we delve into the details of the methodology used in our proposed framework, Anything-3D. Our framework seamlessly integrates visual-language models and object segmentation techniques to address the challenging problem of single-image object reconstruction in unstructured settings. By leveraging the unique strengths of each component, we have developed a powerful approach that can effectively handle the vast variation of images encountered in the wild, across arbitrary categories. Our overall pipeline is shown in Figure~\ref{fig:pipeline}.

\subsection{Identify Object with the Segment-Anything}
\label{subsec:sam}

The first step in our framework is to employ the Segment-Anything Model~(SAM)~\cite{kirillov2023segany} to identify and extract the object regions of interest in the input image. Specifically, SAM takes as input the image $I$ and a prompt $p$ specifying a point or bounding box annotation, and produces accurate segmentation masks $\mathbf{M}$ that mask out the background, later an affine transformation is applied to generate the local image patch $I'$ with the bounding box:

\begin{equation}
\mathbf{M} = \text{SAM}(I, p);
I' = \text{Affine}(\mathbf{M} \odot I; b)
\end{equation}

where $\odot$ denotes the element-wise multiplication using image masking.
SAM is a state-of-the-art object segmentation model that can handle a wide variety of object classes, addressing the challenge of object diversity in the wild.

% \subsection{Image-to-Text Conversion with }
\subsection{Image-to-Text Semantic Conversion}
\label{subsec:blip}

Following the region of interest extraction, we employ a Bootstrapping Language-Image Pre-training (BLIP) model~\cite{li2022blip} to acquire a comprehensive textual description of the input image. The BLIP model, pre-trained on vast quantities of image-text data pairs, develops a profound understanding of object properties, enabling it to provide a reliable description of the target object. Utilizing an image patch $I'$, the BLIP model generates a textual description $T$:

\begin{equation}
T = \text{BLIP}(I')
\end{equation}

Although the extracted $T$ serves as an initial approximation of the image semantics, it may not encompass the precise object information. Consequently, we employ $T$ as the starting point and apply textual inversion~\cite{gal2022image} to estimate the exact soft prompt embedding. Specifically, the initial prompt embedding $e_0 = \textsc{embedding}(T)$ is optimized by matching the score function predicted by the text-to-image diffusion model:

\begin{equation}
e^* = \argmin_{e}||\epsilon_{\phi}(I_t, e) - \bm \epsilon ||_2^2
\end{equation}

Here, $\epsilon_{\phi}$ represents the pre-trained diffusion model~\cite{sdm}, and $I_t$ denotes the noisy image at step $t$:

\begin{align}
I_t = \sqrt{\bar{\alpha}_t} I' + \sqrt{1-\bar{\alpha}_t} \bm{\epsilon}, \quad \text{where} \quad \bm \epsilon \sim \mathcal{N}(0, \mathbf{I}) 
\label{eq:dpm}
\end{align}

The resulting textual embedding $e^*$ serves as invaluable auxiliary information for subsequent 3D reconstruction stages, harnessing the text-image consistency inherent in diffusion models to facilitate a more accurate and comprehensive understanding of the object.

\subsection{3D Reconstruction though Diffusion Prior}
\label{subsec:reconstruction}

With the segmented object and its textual embedding as input, we proceed to the 3D reconstruction task through a coarse-to-fine strategy, as described in the 3DFuse~\cite{3dfuse}. Our 3D reconstruction model leverages an image-to-point cloud and a depth-aware pretrained 2D text-to-image diffusion model, to perform image-to-3D synthesis.

\noindent\textbf{Coarse Geometric Recovery.}
Upon receiving the segmented image $I'$, we exploit the Point-E model~\cite{point_e} to reveal a preliminary estimation of the object's complex structure.  By employing these 3D priors, we meticulously generate a sparse point cloud, which is subsequently projected to a sparse depth map $P$ according to the camera pose $\pi$:

\begin{equation}
P = \mathcal{P}(\text{Point-E}(I'),\pi)
\end{equation}
In this case, $\mathcal{P}(\cdot)$ represents a depth-projection function. This process yields a coarse and sparse depth map, which serves as a geometric constraint, providing crucial guidance for object recovery and laying the foundation for subsequent refinement.

\begin{figure*}
    \centering
    \resizebox{\linewidth}{!}{\includegraphics[width=\linewidth]{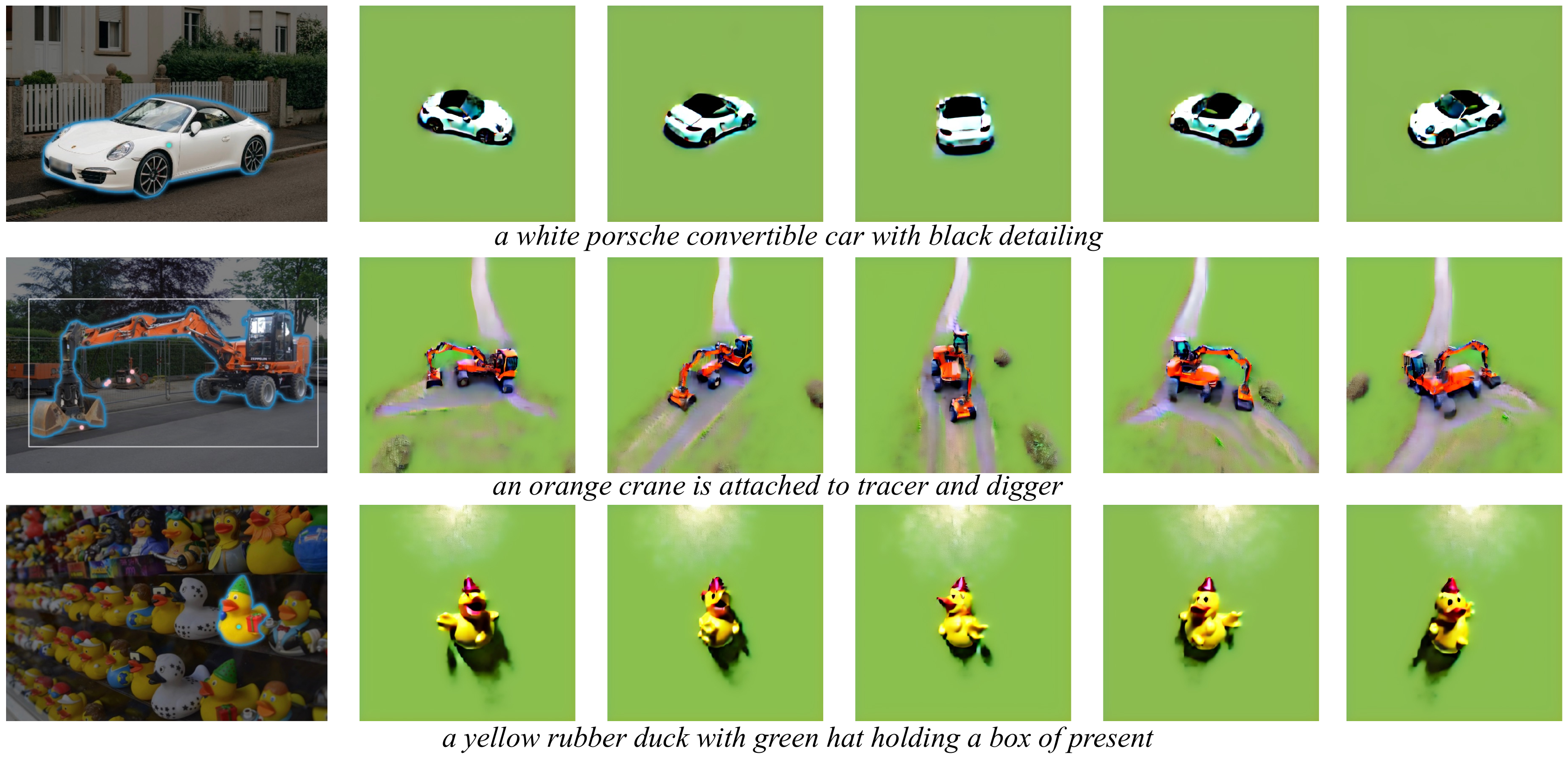}}
    \caption{Generated 3D objects of a convertible car, crane and rubble duck, visualized from five viewpoints.}
    \label{fig:demo1}
    % \vspace{-2mm}
\end{figure*}

\begin{figure*}
    \centering
    \resizebox{\linewidth}{!}{\includegraphics[width=\linewidth]{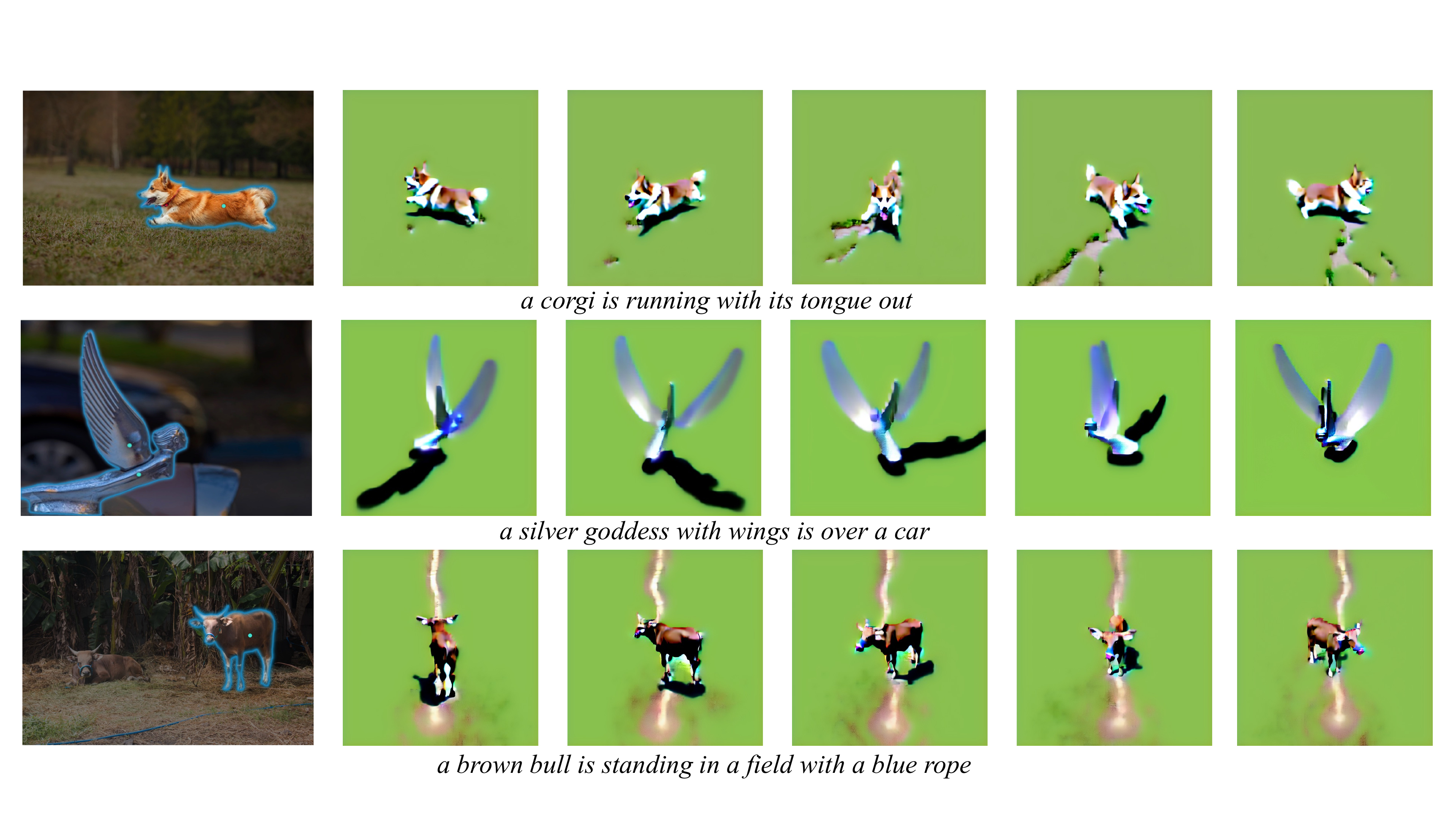}}
    \caption{Generated 3D objects of a cannon, pig bank and stool, visualized from five viewpoints.}
    \label{fig:demo2}
\end{figure*}

\noindent\textbf{Refined 3D Reconstruction.}
Armed with the coarse geometry constraint $P$, the image $I'$, and the text embedding $e^{*}$, we employ an off-the-self text-to-image diffusion models $\epsilon_\phi$, namely Stable-Diffusion~\cite{rombach2022high}, as a prior to synthesize images from alternative camera views, thereby facilitating the estimation of 3D textures from a single image.

In particular, we employ the sparse depth encoder $E_\rho(\cdot)$~\cite{3dfuse}, which ingests the sparse depth map $P$ to control the pose of the generated images. The output features $E_\rho(P)$ are then integrated into the intermediate features of diffusion U-net architecture with ControlNet~\cite{controlnet}. Consequently, the depth map operates as a guiding signal, ensuring that the synthesized images adhere to a consistent geometric perspective.

Afterward, we apply score distillation~\cite{dreamfusion,sjc} to optimize NeRF. In this technique, the parameters of NeRF are updated to follow the score function predicted by the diffusion model. To be more specific, we denote $\theta$ as the parameters of NeRF, and $R_\theta(\pi)$ as the rendering function that takes in a camera pose $\pi$ and produces a rendered image. We use the diffusion model to infer the 2D score of the rendered image, which is then used to optimize the $\theta$. By disregarding negligible terms in the gradient, we obtain the score distillation gradient concerning the NeRF parameters: 

\begin{equation}
    \begin{split}
        \nabla_\Theta \text{L}_{SDS}&(\theta, x =  R_\theta(\pi))= \\ & \mathbb{E}_{t, \rho} \Big[\tilde{w}(t) \big(\epsilon_\phi (x_{t},E_\rho(P),e^*) - \bm\epsilon \big)\frac{\partial x}{\partial\theta} \Big]
    \end{split}
\end{equation}

where $\epsilon_\phi (x_{t}, E_\rho(P),e^*)$ represents  the diffusion model, which accepts the coarse geometry $P$, prompt embedding $e^*$ and noisy rendered image $x_{t}$ as inputs and estimates the score. $x_{t}$ is computed similar to Eq.~\ref{eq:dpm}. For an in-depth derivation, please refer to DreamFussion~\cite{dreamfusion} and SJC~\cite{sjc}.

In conclusion, the optimized NeRF model facilitates the production of a superior 3D reconstruction of the object, derived from a single view image.

\section{Experiment}
As preliminary study, we exam the feasibility of our solution through a series of qualitative visualizations.

\noindent\textbf{Implementation Details.} In practice, we employ a pretrained depth encoder on the Co3D~\cite{co3d} dataset within the 3Dfuse~\cite{3dfuse} framework. We choose a voxel-based Neural Radiance Field~\cite{directvego} to optimize the geometry and texture of 3D objects efficiently. BLIP-2~\cite{li2022blip} is employed to generate the text prompt taking the masked out image patch as input. And a pre-trained Point-E~\cite{point_e} is adopted to initialize a coarse 3D model for generating sparse depth map. We then perform evaluations on selected images from the SA-1B~\cite{kirillov2023segany} dataset, which comprises 11 million images collected from uncontrolled environments.

\noindent\textbf{Results.} The segmentation results and generated 3D objects are presented in Fig.~\ref{fig:demo_0}, Fig.~\ref{fig:demo1} and Fig.~\ref{fig:demo2}. Specifically, we demonstrate the robustness of our framework by reconstructing objects under various challenging conditions, such as occlusion, varying lighting, and different viewpoints. Our framework demonstrates robustness by reconstructing objects under various challenging conditions, such as occlusion, fluctuating lighting, and diverse viewpoints. For instance, we successfully reconstruct irregularly-shaped objects like \emph{crane} and \emph{cannon}. Our model also exhibits competence in handling small objects within cluttered environments, such as the \emph{rubber duck} and \emph{piggy bank}.

These results suggest that our method surpasses existing approaches in terms of accuracy and generalizability, highlighting the effectiveness of our framework in reconstructing 3D objects from single-view images under diverse conditions.

\section{Conclusion}
In this paper, we introduce a novel task aimed at constructing arbitrary 3D objects from single-view, in-the-wild images. We commence by delineating the primary challenges that accompany this task. To surmount these challenges, we devise a versatile framework named Anything-3D. Integrating the zero-shot ability of series of vision and multimodal foundation models, our framework interactively identifies regions of interest within single-view images and represents 2D objects with optimized text embeddings. Ultimately, we efficiently generate high-quality 3D objects using a 3D-aware score distillation model.
In summary, Anything-3D showcases the potential for reconstructing in-the-wild 3D objects from single-view images. We hope our work serves as a solid step towards establishing an agile paradigm for 3D content creation.

\section{Limitations and Future Work}
This preliminary technical report on Single-view Anything Reconstruction in the Wild presents the initial findings and insights of our approach. The reconstruction quality is not yet perfect, and we are continuously working to enhance it. We have not provided quantitative evaluations on 3D datasets, such as novel view synthesis and reconstruction error, but we plan to incorporate these assessments in future iterations of our work.

Furthermore, we aim to expand this framework to accommodate more practical setups, including sparse view object recovery. By addressing these limitations and broadening the scope of our approach, we aspire to refine and augment the performance of our Anything-3D framework.

\clearpage

\newpage
%%%%%%%%% REFERENCES
{\small
\bibliographystyle{ieee_fullname}
\bibliography{egbib}
}

\end{document}